\documentclass[10pt,twocolumn,letterpaper]{article}
 \pdfoutput=1
\usepackage{cvpr}
\usepackage{times}
\usepackage{epsfig}
\usepackage{graphicx}
\usepackage{amsmath}
\usepackage{amssymb}
\usepackage{subcaption}

\usepackage[pagebackref=true,breaklinks=true,letterpaper=true,colorlinks,bookmarks=false]{hyperref}

\cvprfinalcopy 

\def\cvprPaperID{410} 

\ifcvprfinal\fi
\begin{document}

\title{Shadow Optimization from Structured Deep Edge Detection}

\author{Li Shen
	\qquad Teck Wee Chua
	\qquad Karianto Leman\\	
	Institute for Infocomm Research}
	

\maketitle

\begin{abstract}   
   Local structures of shadow boundaries as well as complex interactions of image regions remain largely unexploited by previous shadow detection approaches. In this paper, we present a novel learning-based framework for shadow region recovery from a single image. We exploit the local structures of shadow edges by using a structured CNN learning framework. We show that using the structured label information in the classification can improve the local consistency of the results and avoid spurious labelling. We further propose and formulate a shadow/bright measure to model the complex interactions among image regions. The shadow and bright measures of each patch are computed from the shadow edges detected in the image. Using the global interaction constraints on patches, we formulate a least-square optimization problem for shadow recovery that can be solved efficiently. Our shadow recovery method achieves state-of-the-art results on the major shadow benchmark databases collected under various conditions.
\end{abstract}

\section{Introduction}

\begin{figure*}
	\centering
	\includegraphics[width = 0.9\textwidth]{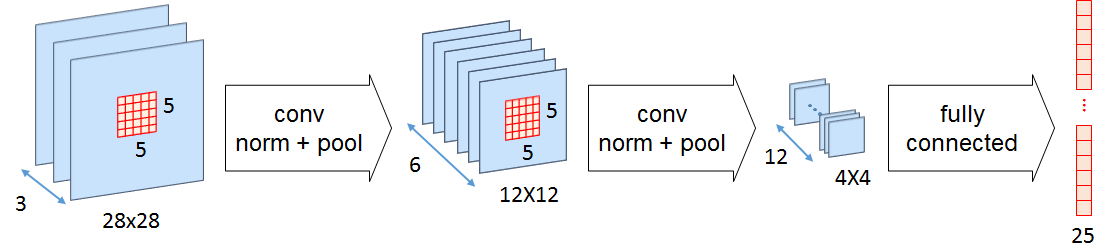}	
	\caption{strCNN architecture used for learning the structure of shadow edges. It takes a $28\times 28$ color image patch as an input, and outputs a ${0, 1}^{25\times 1}$  vector which is corresponding to the shadow edge structure of the $5\times 5$ central patch.}
	\label{fig:CNN}
\end{figure*}

Shadow detection has long been considered a crucial component of scene interpretation. Shadows in an image provide useful information about the scenes: the object shapes~\cite{Okabe09}, the relative positions and 3D structures of the scene~\cite{Caspi06,Abrams14}, the camera parameters and geo-location cues~\cite{Junejo2008}, and the characteristics of the light sources~\cite{Sato03,Lalonde09,Panagopoulos2009}. However, they can also cause great difficulties to many computer vision algorithms, such as background subtraction, segmentation, tracking and object recognition.
Despite its importance and long tradition, shadow detection remains an extremely challenging problem, particularly from a single image. 
The main difficulty is due to the complex interactions of geometry, albedo, and illumination in nature scenes. Since shadows correspond to a variety of visual phenomena, finding a unified approach to shadow detection is difficult. 

Motivated by this observation, recent papers~\cite{Joshi08,Zhu2010,Lalonde2010,Huang2011,Guo2011,Khan2014} have explored the use of learning techniques for shadow detection. These approaches take an image patch and compute the likelihood that the centre pixel contains a shadow edge. Such a classifier is limited by its locality since it treats each pixel independently. 
Optionally, the independent shadow predictions may then be combined using global reasoning by using a CRF/GBP/MRF algorithm. 

Shadow edges in a local patch are actually highly interdependent, and exhibit common forms of local structures: straight lines, corners, curves, parallel lines; while structures such as T-junctions or Y-junctions are highly unlikely on shadow boundaries~\cite{Huang2011,Lalonde2010}.

In this paper we propose a novel learning-based framework for shadow detection from a single image. We exploit the local structures of shadow edges by using a \emph{structured} Convolutional Neural Networks (CNN) framework. A CNN learning framework is designed to capture the local structure information of shadow edges and automatically learn the most relevant features. We formulate the problem of shadow edge detection as predicting local shadow edge structures given input image patches. In contrast to unary classification, we take structured labelling information of the label neighbourhood into account. We show that using the structured label information in the classification can improve the local consistency of the results and avoid spurious labelling.

We also propose a novel global shadow optimization framework. 
In the previous learning approaches, a CRF/GBP/MRF algorithm is usually employed for enforcing the local consistency over neighbouring labels~\cite{Zhu2010,Lalonde2010,Khan2014,Guo2011} and the non-local constraints of region pairs with the same materials~\cite{Guo2011}. The size of the label images and the presence of loops in such a algorithm make it intractable to compute the expectation computations. Moreover, the memory requirement for loading all the training data is large and parameter updating requires tremendous computing resources.
Here, we introduce novel shadow and bright measures to model the region interactions based on the spatial layout of image regions. 
For each image patch, a shadow and a bright measure are computed according to its connectivities to all of the shadow and bright boundaries in the image, respectively. The shadow/bright boundaries are extracted from the shadow edges detected by the proposed CNN. Using these shadow and bright measures, we formulate a least-square optimization problem for shadow recovery to solve for the shadow map(locations). Our optimization framework combines the non-local cues of region interactions in a straightforward and efficient manner where all constraints are in linear form. 

Experimental results on the major shadow benchmark databases demonstrate the effectiveness of the proposed technique. 



\subsection{Related Work}

 Early works for detecting shadows are motivated by physical models of illumination and color~\cite{Jiang1992,Prati2001}. Finlayson et al.~\cite{Finlayson2002,Finlayson2004} located shadows by using a 1-d shadow free image computed from a single image. Their methods can only work under the assumption of approximately Planckian lighting, and high-quality input images are needed. 
 
To adapt to environment changes, statistical learning based approaches~\cite{Porikli05,Liu2007,Huang2009,Joshi08} have been developed to learn the shadow model at each pixel from a video sequence. Recently, some data-driven learning approaches have been developed for single-image shadow detection. Lalonde et.al~\cite{Lalonde2010} detected cast shadow edges on the ground with a classifier trained on local features. \cite{Zhu2010} proposed a similar approach for monochromatic images. Every pixel is classified as either being inside a shadow or not. The per-pixel outputs are inherently noisy with poor contour continuity. To overcome this, the predicted posteriors are usually fed to a CRF formulation which defines the pairwise smoothness constraints across neighbouring pixels. Guo et al.~\cite{Guo2011} modelled the long-range interactions using the non-local cues of region pairs. They then incorporated the non-local pairwise constraints into a graph-cut optimization. Yago et al.~\cite{Yago13} integrated the region, boundary and paired regions classifiers using a MRF. All these learning approaches employ hand-crafted features as input. 

More recently, ~\cite{Khan2014} proposed a deep learning framework to automatically learn the features for shadow detection. They showed that the CNN with learned features outperforms the current state-of-the-art with hand-crafted features. They trained a unary classifier where separated CNNs learned features for detecting shadows at boundaries and uniform regions, respectively. The per-pixel predictions are then fed to a CRF for enforcing local consistency. In contrast to unary classification, we predict the structure of shadow edges in a local patch.
Our work is inspired by the recent works on learning structured labels for semantic image labelling~\cite{Kontschieder11} and edge detection~\cite{Lim13,Dollar13} in Random Forests. Our aim is to explore the structured learning in CNN for local shadow edge detection.

We are also inspired by the work on saliency estimation~\cite{Zhu14}. They propose a background measure based on the boundary connectivity prior that a salient region is less likely connected to the image boundary. Utilizing the connectivity definition, we derive our shadow measures to model the region interactions based on the spatial layout of image regions. 

\section{Structured Deep Shadow Edge Detection}

In this section, we present our structured CNN learning framework, then we explain how to apply it to shadow edge detection. 

\subsection{Structured Convolutional Neural Networks}

Convolutional neural networks actually can have high dimensional and complex output which makes structured output prediction possible. We use $x\in X=\mathbb{R}^{d\times d \times K}$ to denote a color image patch, and $y\in Y= \mathbb{R}^{s \times s}$ to denote the target structured label, where $d$ and $s$ indicate the patch widths of $x$ and the $y$, respectively. $K$ is the number of channels.  
The structured prediction can be expressed as $C: X\rightarrow Y$ mapping from an input domain $X$ to a structured output domain $Y$. Here, $C(\cdot)$ is the structured CNN.   

Fig.~\ref{fig:CNN} shows our network architecture with 7-layers. 
Our learning approach predicts a structured $y^{25\times 1}$ label from a larger  $28\times 28$ image patch.
The network consists of two alternating convolutional and max-pooling layers, followed by a fully connected layer and finally a logistic regression output layer with “softmax” nonlinear function.
The first and second convolutional layers consist of six and twelve $5\times5$ kernels, respectively, with unit pixel stride. The pooling size is $3\times3$ with unit-pixel stride.  Sequentially, the fully connected layer has 64 hidden units. 
 
One challenge for training CNN with structured labels is that structured output spaces are high dimensional, which causes long training duration. We show in the experiments that with the setting of $y^{5\times5}$ we can capture wide variety of local shadow structures sufficiently while still keeping the training complexity low.

\subsection{Structured Learning Shadow Edges}

We employ the proposed structured CNN for feature learning for shadow edge detection. It takes a $28\times 28$ color image patch as an input, and outputs a $\{1,0\}^{25\times1}$ vector which is corresponding to the $5 \times 5$ shadow probability map of the central patch.

Assume we have a set of images $X$ with a corresponding set of binary images $Y$ of shadow edges. The structured CNN operates on the patches at image edges: only the patches that contain image edges at its $5\times5$ central area are used. The input patches $\{x^{28 \times 28}\}$ are extracted from $X$. The corresponding groundtruth $\{y^{5\times5}\}$ are extracted from the binary images $Y$. We randomly sample $N(<800)$  shadow and non-shadow edge patches, respectively, per image for training. Before feeding the extracted patches to the CNN, the data is zero-centered and normalized. 

Specifically, we first apply Canny edge detector to extract all shadows and non-shadows. 
Positive patches are obtained from the pixels on Canny edges that coincide with the groundtruth shadow edges. 
Likewise, negative (non-shadow) input patches are obtained from the edge pixels that do not overlap with the groundtruth shadow edges. 
In the experiments, as the groundtruth is hand-labelled by human, the actual shadow edge may not coincide with the groundtruth edges. We usually dilate the groundtruth edges before overlap with the Canny edges. For the datasets with region-based groundtruth such as UCF and UIUC, we extract the blob boundaries as the groundtruth of shadow edges. 
As pointed out in~\cite{Khan2014} non-shadow pixels usually outnumber shadow pixels by approximately 6:1 ratio. We address this class imbalance problem by setting the number of positive samples as the upper bound of the number of negative samples to be sampled. To reduce the number of redundant samples, we only randomly sample the patches on $3\times3$ grid. During the training process, we use stochastic gradient descent.

\begin{figure}
	\centering
	\begin{subfigure}[b]{0.21\textwidth}
		\includegraphics[width=\textwidth]{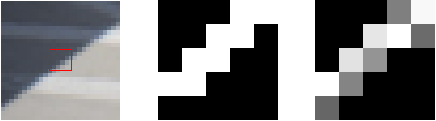}\vspace{2pt}
		\includegraphics[width=\textwidth]{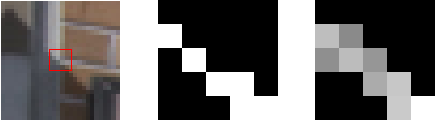}\vspace{2pt}
		\includegraphics[width=\textwidth]{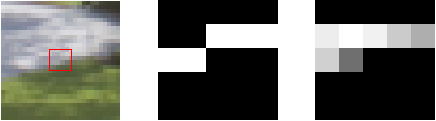}\vspace{2pt}
		\includegraphics[width=\textwidth]{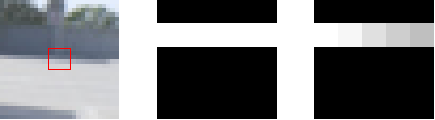}\vspace{2pt}
		\includegraphics[width=\textwidth]{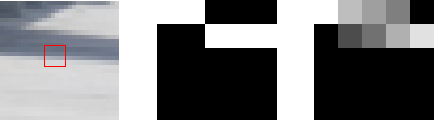}\vspace{2pt}		
		\caption{}
		\label{fig:structuredlabel_shadow}
	\end{subfigure}%
	\qquad 
	\begin{subfigure}[b]{0.21\textwidth}
		\includegraphics[width=\textwidth]{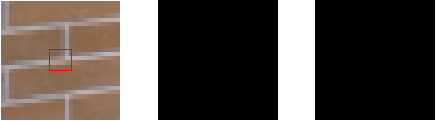}\vspace{2pt}
		\includegraphics[width=\textwidth]{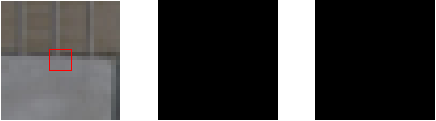}\vspace{2pt}
		\includegraphics[width=\textwidth]{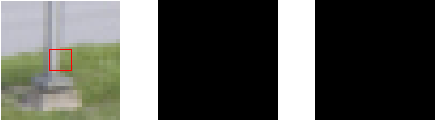}\vspace{2pt}
		\includegraphics[width=\textwidth]{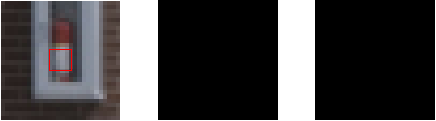}\vspace{2pt}
		\includegraphics[width=\textwidth]{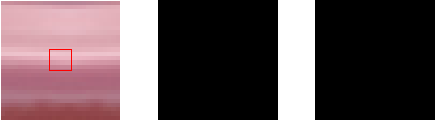}\vspace{2pt}
		\caption{}
		\label{fig:structuredlabel_light}
	\end{subfigure}	
	\caption{(a) Shadow and (b) non-shadow patches learned by the proposed structure CNN. Left: input $28\times28$ patches(red rectangle indicates the central $5\times5$ region). Center: groundtruth $5\times5$ patches. Right: output $5\times5$ patches.}\label{fig:structuredlabel}
\end{figure}

Fig.~\ref{fig:structuredlabel} shows the shadow and non-shadow edges learned by the proposed structure CNN. We can see that our CNN does capture the local structures of shadow edges. Besides learning the classification of shadow and non-shadow edges, the proposed network also learns valid labelling transitions among adjacent pixels (i.e. interactions among the pixels in a local patch).

Our CNN was implemented in unoptimized Matlab code. The training took $\sim 5$ hours on $5 \times 5$ pixels structured output ($\sim 700$ iteration) given input size of $500K+$ patches and consumed $\sim 6GB$ memory on Intel Quad-Core $2.4GHz$ PC. 

\subsection{Structured Labelling Shadow Edges}

Given an input image, Canny edge detector is applied to find the significant edges in the image. Then, we extracted windows along the image edges. The overlapping edge patches are then fed to the proposed CNN for labelling. The trained structured CNN differentiates between the shadow and reflectance edges and predicts the shadow edge structure of the central region. Our structured CNN achieves robust results by combining the predictions of the neighboring ones. Instead of independently assigning a class label to each pixel, our structured labels predict the interactions among the neighbouring pixels of a local patch. Each pixel collects class hypotheses from the structured labels predicted for itself and neighboring pixels. We employ a simple voting scheme to combine the multiple predicts at each pixel at the image edges. 

\begin{figure*}
	\begin{center}
		\begin{subfigure}[b]{0.27\textwidth}
			\includegraphics[width=\textwidth, height = 3.6cm]{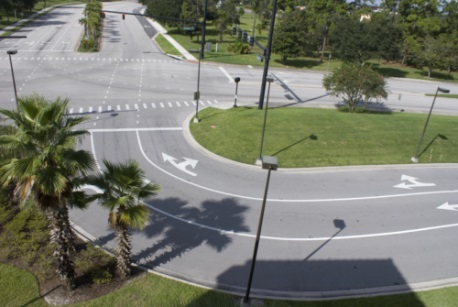}\vspace{2pt}
			\includegraphics[width=\textwidth, height = 3.6cm]{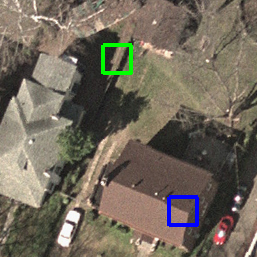} 
			\caption{}
			\label{fig:cnn_org}
		\end{subfigure}%
		\quad 
		\begin{subfigure}[b]{0.27\textwidth}
			\includegraphics[width=\textwidth, height = 3.6cm]{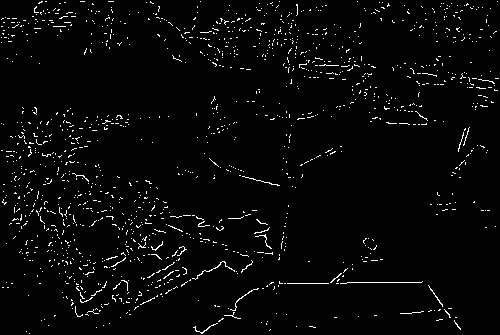}\vspace{2pt}
			\includegraphics[width=\textwidth, height = 3.6cm]{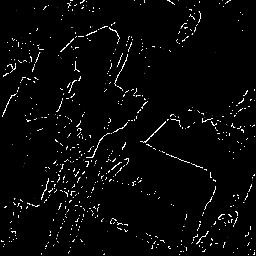} 
			\caption{}
			\label{fig:cnn_1x1}
		\end{subfigure}
		~
		\begin{subfigure}[b]{0.27\textwidth}
			\includegraphics[width=\textwidth, height = 3.6cm]{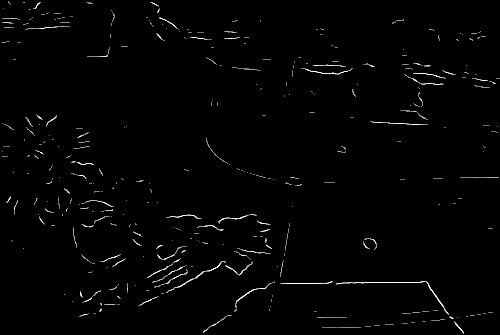}\vspace{2pt}
			\includegraphics[width=\textwidth, height = 3.6cm]{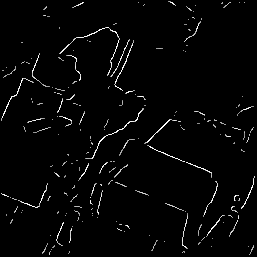} 
			\caption{}
			\label{fig:cnn_5x5}
		\end{subfigure}
		~
		\begin{subfigure}[b]{0.11\textwidth}
			\includegraphics[width=1.5cm, height = 3.6cm]{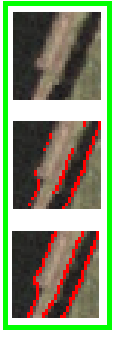} \vspace{2pt}
			\includegraphics[width=1.5cm, height = 3.6cm]{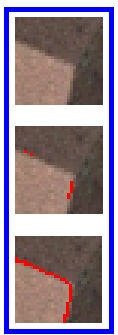} 
			\caption{}
		\end{subfigure}	
	\end{center}
	\caption{Structured shadow edge detection results. (a) input image. We compare the detection results using (b) 1x1 labelling CNN(previous method), (c) 5x5 structured labelling CNN. (d) Zoom-in of green and blue patches. Top to bottom:original, 1x1 and 5x5 outputs. We can see that 5x5 structured CNN is able to learn fine shadow details. 1x1 output has serious spurious noise.}\label{fig:cnn1x1_cnn5x5}
\end{figure*}

Fig.~\ref{fig:cnn1x1_cnn5x5} illustrates the advantage of shadow edge detection with structured output CNN. As can be seen, the
proposed structured CNN can recover better local edge structures (local consistency), avoid assigning implausible label transitions.

\section{Shadow optimization}

We first derive the local and global shadow/bright measures to model the interactions among the regions across the image. Then, we present our optimization framework to solve for the shadow map.  

\subsection{Global and local shadow(bright) probability}

\begin{figure}
	\begin{center}
		\includegraphics[width = 2.1in, height= 1.6in]{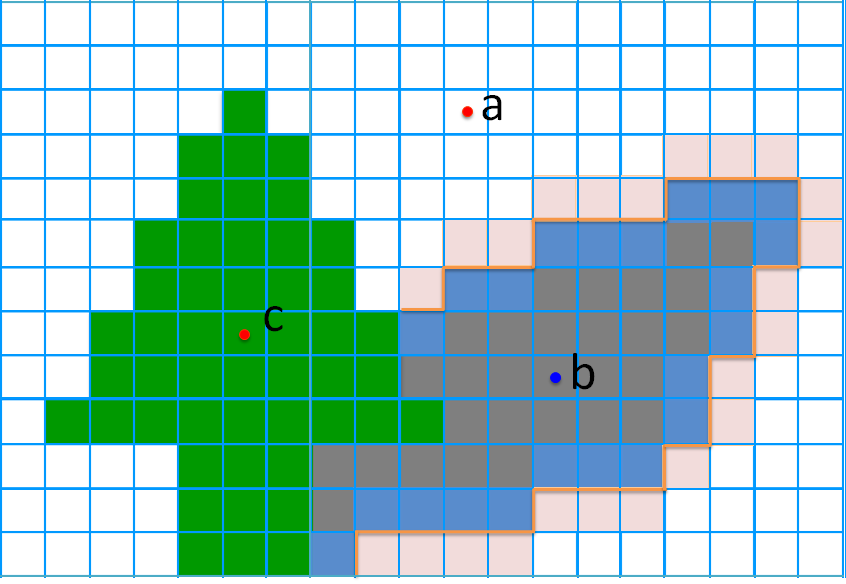}
	\end{center}
	\caption{ An illustrative example. The orange lines are shadow edges. The blue and pink regions adjacent to the shadow edges are the shadow and bright boundaries, respectively. }
	\label{fig:shadowlightlayout}
\end{figure}

We observe that both shadow and bright regions have the following characteristic in their spatial layout: shadow regions are much more connected to shadow boundaries, and bright regions are more connected to bright boundaries. We define the dark side region of a shadow edge as the \emph{shadow boundary}, while the bright side region as the \emph{bright boundary}.  In Fig.~\ref{fig:shadowlightlayout}, we illustrate a tree and its shadow. The blue and pink regions are the shadow and bright boundaries, respectively. The grey region is clearly a shadow region as it significantly touches the shadow boundary, while the white region is clearly a bright region as it largely touches the bright boundary. 

The geodesic distance between any two patches $p$, $q$ in an image is defined as the accumulated edge weights along their shortest path on the graph:
\begin{equation*}
d_{geo}(p, q) = min_{p1=p, p_2, ..., p_n=q} \sum_{i=1}^{n-1}d_{app}(p_i, p_{i+1})
\end{equation*}
where $d_{app}(p, q)$ is the Euclidean distance between the average colours of the two patches. The normalized $D(p, q) \in (0, 1]$ is thus defined as $D(p, q) = exp(-\frac{d^2_{geo}(p, q)}{2\sigma_{clr}^2})$ which characterizes how much patch $p$ connects(or contributes) to patch $q$. $D(p, q)\approx 0$ when $d^2_{geo}(p, q)\gg \sigma_{clr}$.

Let $\mathbf{shd}$ be the set of shadow boundary patches, and $\mathbf{lit}$ be the set of bright boundary patches. Following the boundary connectivity introduced in~\cite{Zhu14}, we formulate the shadow and bright boundary connectivities of a patch $p$ as: 
 \begin{equation}
 	 con_{shd}(p) = \frac{Len(p, \mathbf{shd})}{\sqrt{Area(p)}} ~\text{and}~ con_{lit}(p) = \frac{Len(p, \mathbf{lit})}{\sqrt{Area(p)}},
 	 \label{equ:Connectivity}
 \end{equation} 
respectively. $Len(p, \mathbf{x})=\sum_{q\in x} D(p, q)$ is the connectivity of $p$ to boundary $\mathbf{x}$. $Area(p) = \sum_{i=1}^N D(p, p_i)$ is the spanning area of $p$, where $N$ is the number of patches in the image.   
Note that we set $d_{app}(p\in\mathbf{shd}, q\in\mathbf{lit})= \inf$, which implies that the bright and shadow boundary sets are not connected, and no path can cut through the shadow edges. We can see that $con_{shd/lit}(p)$ quantifies how heavily a patch $p$ is connected to the shadow/bright boundaries in a local area ($\sigma_{clr} = 5$ in the experiments.) 

Hence, we define the local shadow/bright measure as:
\begin{equation}
\gamma_{shd,lit}(p) = exp(-\frac{con_{shd,lit}(p)}{2\sigma_{con}^2}),
\label{equ:localweight}
\end{equation}
and local shadow/bright probability can be computed as $Pr_{loc}^{shd,lit}(p) = 1-\gamma_{shd,lit}(p)$ which is close to 1 when shadow/bright connectivity is large, and 0 when it is small. Note that $Pr_{loc}(p)$ is only affected by the local shadow edges in the region $p$ belong to. If a patch is hardly connected to  all the shadow edges in the image, $Pr_{loc}(p)$ is low, which indicates that we cannot get a correct prediction with the local information. 

We define the global shadow/bright measure as:
\begin{equation}
 \Gamma_{shd, lit}(p) = \sum_{i=1}^N w_{app}(p,p_i)w_{spa}(p, p_i)\gamma_{shd,lit}(p)
 \label{equ:globalweight}
\end{equation}
where $w_{app} = exp(-\frac{d_{app}^2}{2\sigma_{app}^2})$, and $w_{spa} = exp(-\frac{d_{spa}^2}{2\sigma_{con}^2})$. $d_{app}$ and $d_{spa}$ are the Euclidean distance between the average colors and locations, respectively. It is based on the observation that if two patches in an image are with the same colour and near to each other, they usually are both in shadow or bright regions. Global shadow/bright probability at $p$ can be computed as $Pr_{glb}^{shd,lit}(p) = 1- \Gamma_{shd, lit}(p)$.  

\begin{figure*}
	\centering
	\begin{subfigure}[b]{0.14\textwidth}
		\includegraphics[width=\textwidth]{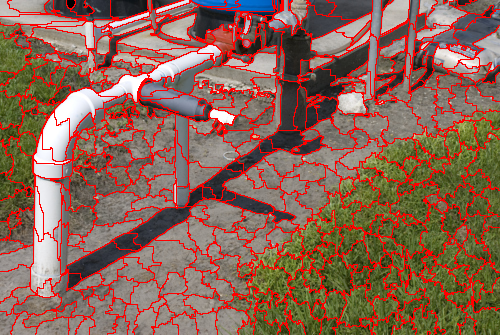} 
		\includegraphics[width=\textwidth]{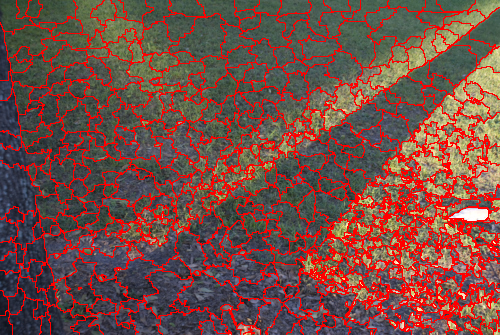} 
		\caption{}
	\end{subfigure}%
	\hfill
	~
	\begin{subfigure}[b]{0.14\textwidth}
		\includegraphics[width=\textwidth]{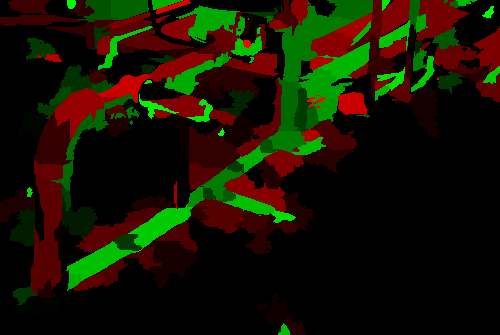} 
		\includegraphics[width=\textwidth]{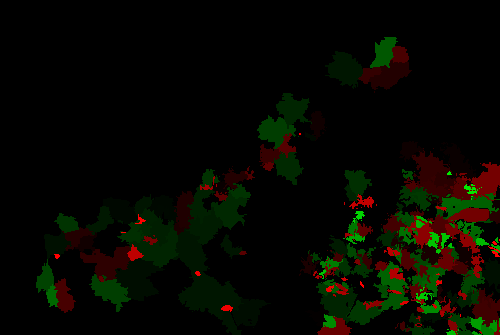}	
		\caption{}
	\end{subfigure}
	\hfill
	\begin{subfigure}[b]{0.28\textwidth}
		\includegraphics[width=\textwidth]{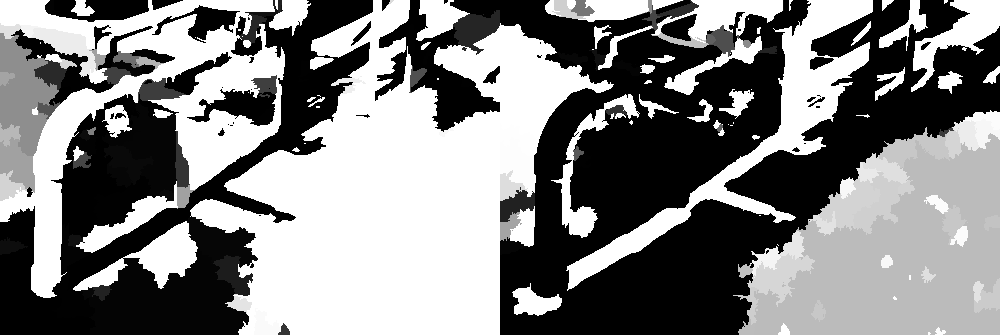} 
		\includegraphics[width=\textwidth]{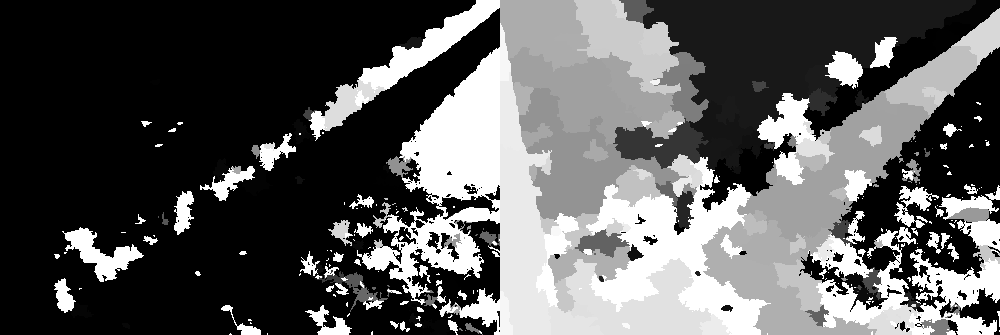}
		\caption{}
	\end{subfigure}
	\hfill
	\begin{subfigure}[b]{0.28\textwidth}
		\includegraphics[width=\textwidth]{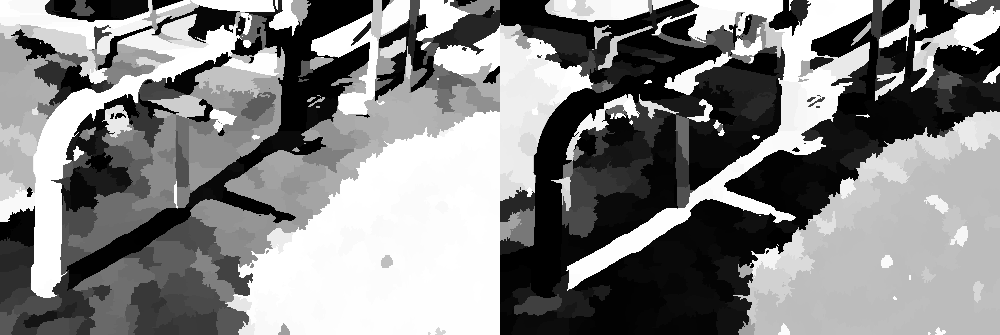} 
		\includegraphics[width=\textwidth]{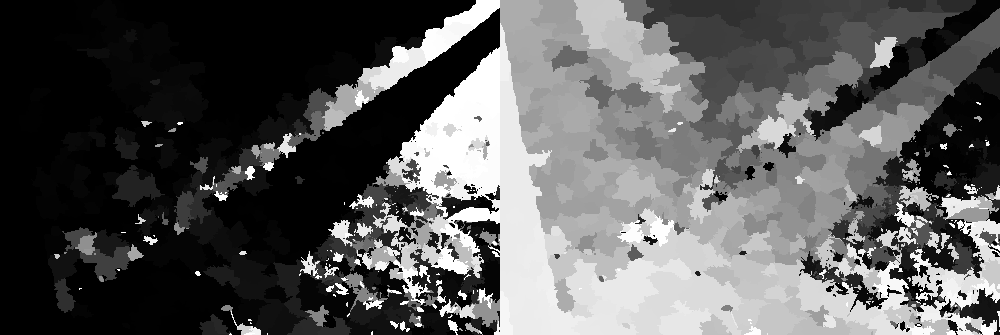}
		\caption{}
	\end{subfigure}
	\hfill
	\begin{subfigure}[b]{0.14\textwidth}
		\includegraphics[width=\textwidth]{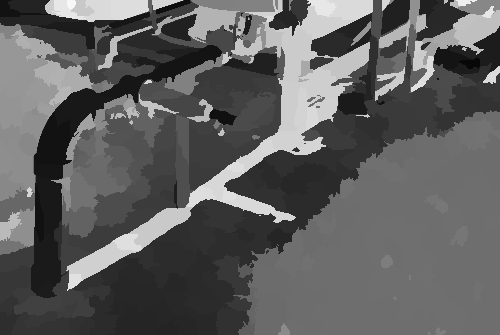} 
		\includegraphics[width=\textwidth]{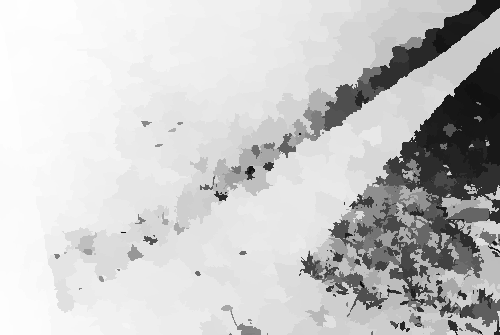}	
		\caption{}
	\end{subfigure}
	\hfill
	\caption{Global shadow optimization. (a) input images with superpixel boundaries overlaid. (b)shadow and bright boundary patches extracted from the local detected shadow edges. (c) local shadow(L) and bright(R) measures in Eq.~\ref{equ:localweight}; (d) global shadow(L) and bright(R) measures in Eq.~\ref{equ:globalweight}. The predicts are propagated over the image. (e) optimized shadow maps by minimizing Eq.~\ref{equ:globalopt}.} 
\end{figure*}

\subsection{Global shadow optimization}

The input image is first abstracted as a set of nearly regular superpixels using the Quick Shift segmentation method~\cite{Vedaldi08}. 
The shadow and bright boundary patches can be extracted from the local shadow edge detection results. We only select the most reliable patches as the shd and lit boundary regions. Specifically, if a superpixel at shadow edge is darker(brighter) than all the adjacent superpixels, we set it  as a shadow(bright) boundary patch. If a superpixel is brighter than some of its neighbours but darker than some others, we consider the patch to be ambiguous and it would be discarded.  

After obtaining $\mathbf{shd}$ and $\mathbf{lit}$, the local shadow/bright measures $\gamma_{shd, lit}$ can be computed at each superpixel respectively, as described in Eq.~\ref{equ:localweight}.  The global shadow/bright measures $\Gamma_{shd, lit}(p)$ can be computed as in Eq.~\ref{equ:globalweight}, consequently,.

We formulate the shadow detection problem as the optimization of the shadow values of all image superpixels. The objective cost function is designed to assign the shadow regions value 1 and the bright regions value 0, respectively.The optimal shadow map is then obtained by minimizing the cost function. Let the shadow values of $N$ superpixels be $\{s_i\}_{i=1}^N$. The cost function is thus defined as
\begin{equation}
\begin{split}
E = &\sum_{i=1}^N w_i^{shd}s_i^2 + \sum_{i=1}^N w_i^{brt}(1-s_i)^2 + \sum_{i,j} w_{ij}(s_i-s_j)^2 \\ &+\lambda\sum_{i\in\mathbf{shd, lit}} (s_i-\tilde{s_i}_)^2
\end{split} 
\label{equ:globalopt}
\end{equation}
where $ w_i^{shd,lit} = \Gamma_{shd, lit}(p_i)$ defined in Eq.~\ref{equ:globalweight}. $\tilde{s_i}$ is the initial values for $p\in \{ \mathbf{lit},  \mathbf{shd}\}$ ,  where $\tilde{s_i} =1$ for $p_i \in \mathbf{shd}$ and 0 for $p_i \in \mathbf{lit}$. We set $\lambda=0.001$ in the experiments.

The four terms are all squared errors and the optimal shadow map is computed by least-square. Fig. 5 shows the optimized results. 

\section{Experiments}

\subsection{Datasets}

\textbf{UCF Shadow Dataset:} This dataset contains $355$ images with manually labeled region-based ground truth. Only $245/355$ images were used in~\cite{Zhu2010, Guo2011}. The split of the train/test data is according to the software package provided by~\cite{Guo2011} as the original authors did not disclose the split.\\
\textbf{CMU Shadow Dataset:} This dataset contains $135$ images with manually labeled edge-based ground truth for shadow on the ground. As our algorithm is not restricted to ground shadows, we create the ground plane masks and augment the edge-based ground truth to region-based ground truth. The authors did not report train/test data split therefore we follow the procedure in~\cite{Khan2014} where even images for training and odd images for testing.\\
\textbf{UIUC Shadow Dataset:} This dataset contains $108$ images ($32$ train images and $76$ test images) with region-based ground truth. 

\subsection{Results}

We extensively evaluated our proposed algorithm on three publicly available single image shadow datasets. The evaluation results on UCF in~\cite{Khan2014} were based on the full dataset. To be comparable to their results, we reported both the UCF results using the full dataset and the subset 245 images. As shown in Table~\ref{tab:OverallAcc}, our shadow detection method(SCNN-LinearOpt) achieves the best performances for all three datasets. In particular,we achieve almost 2\% and 5\% gain over state-of-the-art results for UCF and CMU datasets. Table~\ref{tab:ClassWiseAcc} shows the comparisons of class-specific detection accuracies. We take the shadow accuracy to be the number of pixels correctly detected as shadow divided by the total number of pixels marked as shadow in the ground truth. Likewise, non-shadow accuracy is obtained in a similar manner. Our approach achieves significantly higher shadow accuracies. This is consistent with the finding from Fig.~\ref{fig:ROC} where our approach delivers highest AUC. 

\begin{figure*}
	\centering
	\includegraphics[width = 7in, height= 3.2in]{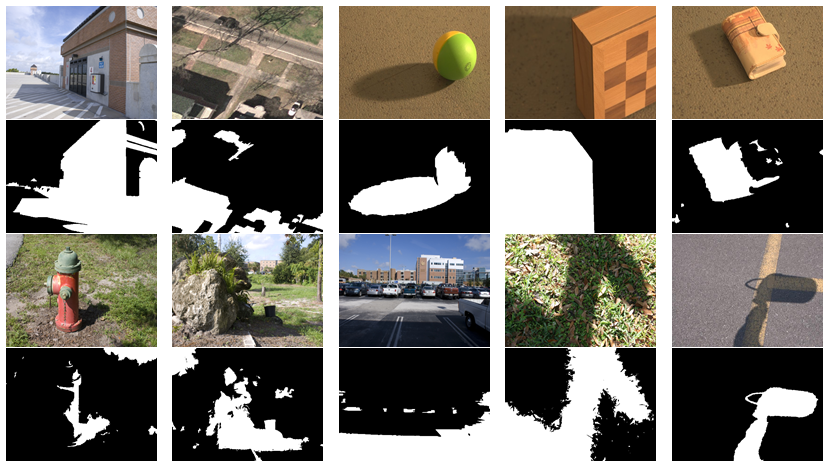}
	\caption{Shadow optimization results from detected shadow edges. Top: input shadow images. Bottom: recovered shadow regions.} \label{fig:Resultimg}
\end{figure*}

\begin{figure*}
	\centering
	\begin{subfigure}[b]{0.32\textwidth}
		\includegraphics[width=\textwidth]{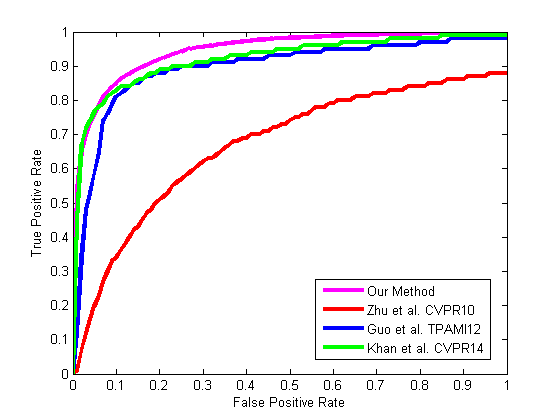}
		\label{fig:ucf_roc}
	\end{subfigure}%
	~ 
	\begin{subfigure}[b]{0.32\textwidth}
		\includegraphics[width=\textwidth]{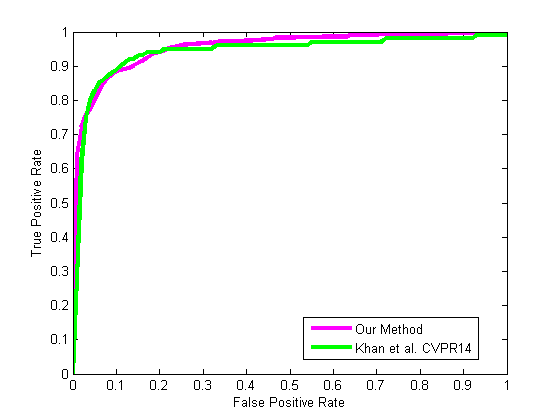}
		\label{fig:ucf_roc}
	\end{subfigure}
	~
	\begin{subfigure}[b]{0.32\textwidth}
		\includegraphics[width=\textwidth]{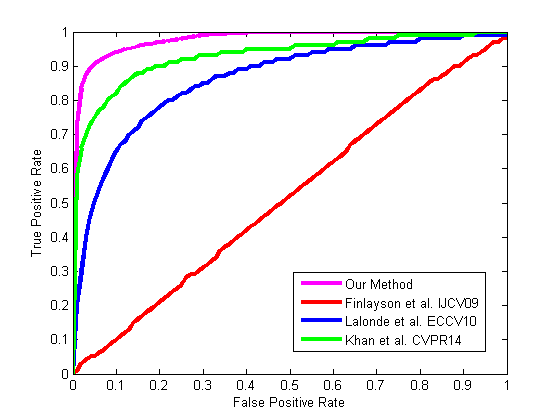}
		\label{fig:ucf_roc}
	\end{subfigure}
	\caption{ROC curves on (a) UCF dataset, (b) UIUC dataset, and (c) CMU dataset}\label{fig:ROC}
\end{figure*}

\begin{figure}
	\centering
	\includegraphics[width = 3.3in]{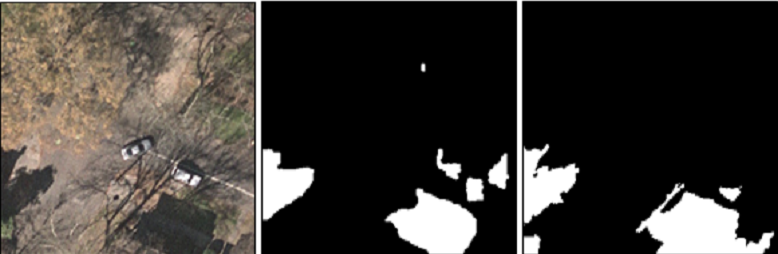}	
	\caption{Comparison with Zhu's work~\cite{Zhu2010}. Left:input image. Middle:Zhu's results. Right: our result.}
	\label{fig:comparison_Zhu}
\end{figure}

\begin{figure}
	\centering	
	\includegraphics[width = 3.3in]{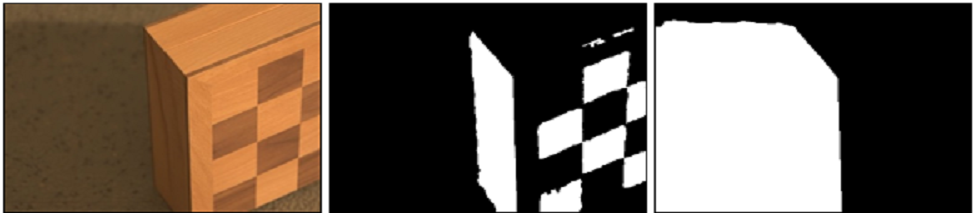}
	\caption{Comparison with Guo's work~\cite{Guo2011}. Left:input image. Middle: Guo's results. Right: our result. Our method can correctly recovery the shadow regions.}
	\label{fig:comparison_guo}
\end{figure}

\begin{figure}
	\begin{center}	
		\begin{subfigure}[b]{0.155\textwidth}
			\includegraphics[width=\textwidth]{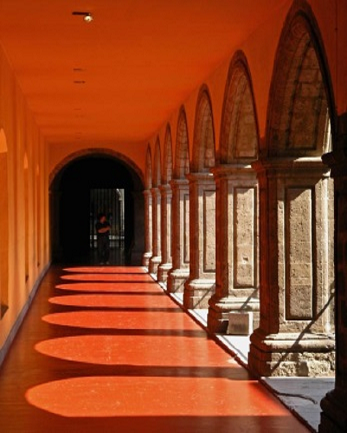}				
		\end{subfigure}%
		~		
		\begin{subfigure}[b]{0.155\textwidth}
			\includegraphics[width=\textwidth]{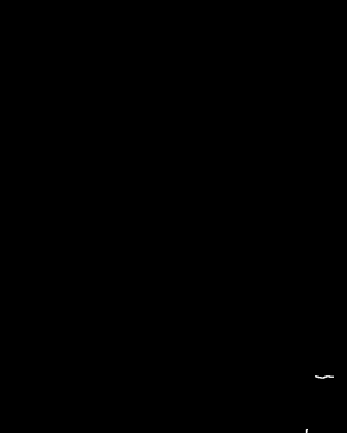}			
		\end{subfigure}%
		~
		\begin{subfigure}[b]{0.155\textwidth}
			\includegraphics[width=\textwidth]{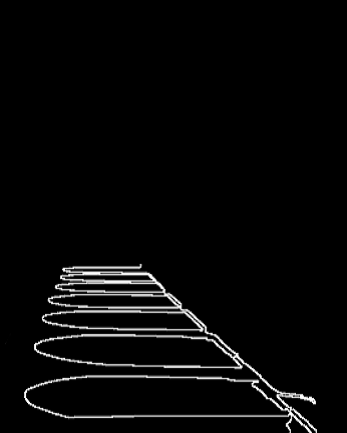}			
		\end{subfigure}%
	\end{center}
	\caption{Shadow edge detection results comparing with Lalonde's work~\cite{Lalonde2010}. Left:input image. Middle: Lanlonde's results. Right: our result. Our method can accurately detect the shadow edges.}
	\label{fig:comparison_lan}
\end{figure}

\begin{figure*}
	\centering
	\includegraphics[width = 6.8in]{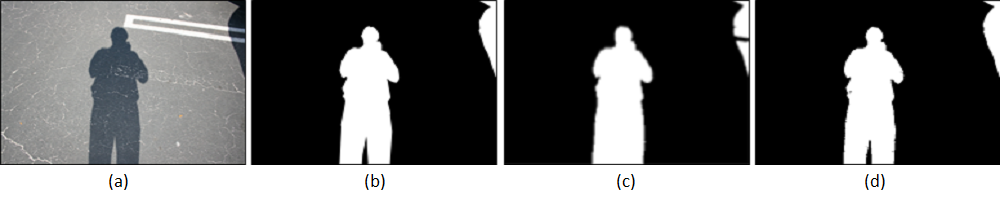}
	\caption{Comparison with Khan's work~\cite{Khan2014}. Left:input image. Middle: Khan's results. Right: our result.}
	\label{fig:comparison_khan}
\end{figure*}

Fig.~\ref{fig:Resultimg} shows some of the qualitative results obtained with our method. The results suggest that our shadow detectors performed robustly under various cases ranging from indoor images to outdoor and aerial images that exhibit soft shadow, low light condition, colour cast, and complex self-shading regions. In Fig.~\ref{fig:comparison_Zhu}, we compare our approach with Zhu’s work~\cite{Zhu2010}. Our method can correctly recover the shadow regions in the complex scene. In Fig.~\ref{fig:comparison_guo}, we show that our approach outperforms Guo's work~\cite{Guo2011} in the ambiguous situation that the object material has the similar colour of the shadows in the image. In Fig.~\ref{fig:comparison_lan}, we compare our shadow edges results with the Lalonde's results~\cite{Lalonde2010}. Our method can accurately detect the shadow edges of the image which Lalonde's method fails with. We also compare our method with Khans et al.’s very recent work~\cite{Khan2014} in Fig.~\ref{fig:comparison_khan}.

\begin{table*}
	\centering
	\begin{tabular}{l|ccc}
		\hline 
		\textbf{Methods} & \textbf{UCF dataset} & \textbf{UIUC dataset} & \textbf{CMU dataset}\tabularnewline
		\hline 
		\hline
		BDT-BCRF~\cite{Zhu2010} & 88.7\% & - & -\tabularnewline
		BDT-CRF-Scene~\cite{Lalonde2010} & - & - & 84.8\%\tabularnewline
		Unary-Pairwise~\cite{Guo2011} & 90.2\% & 89.1\% & -\tabularnewline
		CNN-CRF~\cite{Khan2014} & 90.7\%* & 93.2\% & 88.8\%\tabularnewline
		SCNN-LinearOpt & \textbf{93.1 \%}(\textbf{92.3\%}*) & \textbf{93.4\%} & \textbf{94.0\%}\tabularnewline
		\hline
	\end{tabular}\caption{Performance Comparisons of Shadow Detection Methods}\label{tab:OverallAcc}
\end{table*}

\begin{table}
\begin{tabular}{lcc}
\hline 
\textbf{Datasets\textbackslash{}Methods} & \textbf{Shadows} & \textbf{Non-Shadow}\tabularnewline
\hline
\hline 
\textbf{UCF Dataset} &  & \tabularnewline
\hline 
BDT-BCRF~\cite{Zhu2010} & 63.9\% & 93.4\%\tabularnewline
Unary-Pairwise~\cite{Guo2011} & 73.3\% & \textbf{93.7\%}\tabularnewline
CNN-CRF~\cite{Khan2014} & 78.0\%* & 92.6\%*\tabularnewline
SCNN-LinearOpt & \textbf{91.1\%}(\textbf{91.6\%}*) & 93.5\%(93.4\%*)\tabularnewline
\hline 
\textbf{UIUC Dataset} &  & \tabularnewline
\hline 
Unary-Pairwise~\cite{Guo2011} & 71.6\% & 95.2\%\tabularnewline
CNN-CRF~\cite{Khan2014} & 84.7\% & \textbf{95.5\%}\tabularnewline
SCNN-LinearOpt & \textbf{91.3\%} & 95.03\%\tabularnewline
\hline 
\textbf{CMU Dataset} &  & \tabularnewline
\hline 
BDT-CRF-Scene~\cite{Lalonde2010} & 73.1\% & 96.4\%\tabularnewline
CNN-CRF~\cite{Khan2014} & 83.3\% & 90.9\%\tabularnewline
SCNN-LinearOpt & \textbf{91.6\%} & \textbf{97.7\%}\tabularnewline
\hline
\end{tabular}\caption{Pixel-wise Shadow/Non-Shadow Detection Accuracy. \footnotesize{*: The result is produced using the full dataset.}}\label{tab:ClassWiseAcc}
\end{table}

\section{Conclusions}
In this paper, we propose an efficient structured labelling framework for shadow detection from a single image. We show that the structured CNN Networks framework can capture the local structure information of shadow edge. Moreover, we present a novel global shadow/bright measures to model the complex global interactions based on spatial layout of image regions. The non-local constraints on shadow/bright regions help to overcome ambiguities in local inference. Using these non-local region constraints, we formulate the shadow detection as a least-square optimization problem which can be solved efficiently. Our method can be easily extended to other low-level problems such as object edge detection, smoke region detection etc..

{\small
\bibliographystyle{ieee}
\bibliography{shadow}
}

\end{document}